\documentclass[conference]{IEEEtran}
\IEEEoverridecommandlockouts
\usepackage{cite}
\usepackage{amsmath,amssymb,amsfonts}
\usepackage{algorithmic}
\usepackage{graphicx}
\usepackage{textcomp}
\usepackage{multirow}
\usepackage{color,soul}
\usepackage[table,xcdraw]{xcolor}
\usepackage[caption=false,font=footnotesize]{subfig}
\def\BibTeX{{\rm B\kern-.05em{\sc i\kern-.025em b}\kern-.08em
    T\kern-.1667em\lower.7ex\hbox{E}\kern-.125emX}}

\makeatletter
\def\ps@IEEEtitlepagestyle{
  \def\@oddfoot{\mycopyrightnotice}
  \def\@evenfoot{}
}
\def\mycopyrightnotice{
  {\parbox{\textwidth}{\footnotesize \copyright~2026 IEEE. This work has been accepted for publication in IEEE DCAS 2026. The final published version will be available via IEEE Xplore Permission from IEEE must be obtained 
  for all other uses, in any current or future media, including 
  reprinting/republishing this material for advertising or promotional 
  purposes, creating new collective works, for resale or redistribution 
  to servers or lists, or reuse of any copyrighted component of this 
  work in other works.}}
  \gdef\mycopyrightnotice{}
}
\makeatother

\begin{document}

\title{Quantum-Enhanced Vision Transformer for Flood Detection using Remote Sensing Imagery\\
}

\author{%
\IEEEauthorblockN{Soumyajit Maity \quad and \quad Behzad Ghanbarian}
\IEEEauthorblockA{%
\textit{iResearchE$^3$ Lab, Department of Earth and Environmental Sciences}\\
\textit{University of Texas at Arlington}, Arlington, TX, USA\\
sxm7770@mavs.uta.edu, ghanbarianb@uta.edu}
}

\maketitle

\begin{abstract}
Reliable flood detection is critical for disaster management, yet classical deep learning models often struggle with the high-dimensional, nonlinear complexities inherent in remote sensing data. To mitigate these limitations, we introduced a novel Quantum-Enhanced Vision Transformer (ViT) that synergizes the global context-awareness of transformers with the expressive feature extraction capabilities of quantum computing. Using remote sensing imagery, we developed a hybrid architecture that processes inputs through parallel pathways, a ViT backbone and a quantum branch utilizing a 4-qubit parameterized quantum circuit for localized feature mapping. These distinct representations were fused to optimize binary classification. Results showed that the proposed hybrid model significantly outperformed a classical ViT baseline, increased overall accuracy from 84.48\% to 94.47\% and the F1-score from 0.841 to 0.944. Notably, the quantum integration substantially improved discriminative power in complex terrains for both class. These findings validate the potential of quantum-classical hybrid models to enhance precision in hydrological monitoring and earth observation applications.
\end{abstract}


\section{Introduction} 

Flooding is among the most devastating natural hazards worldwide, responsible for substantial loss of life and economic damage amounting to tens of billions of dollars annually. The increasing frequency and intensity of extreme precipitation events, driven in part by climate change and rapid urbanization, have further exacerbated flood risks across both developed and developing regions. Consequently, reliable and timely flood detection has become a critical component of disaster risk reduction, emergency response, and long-term infrastructure planning. Accurate flood detection enables authorities and stakeholders to issue early warnings, allocate resources effectively, and implement mitigation strategies that can significantly reduce human and economic losses.

Despite its importance, flood detection remains a complex and challenging task. It requires high-quality hydrological and geospatial data describing the physical and environmental characteristics of a given area, including topography, land use and land cover, soil properties, drainage networks, and meteorological forcing. These variables interact in highly nonlinear ways across multiple spatial and temporal scales, making the underlying processes difficult to model using purely physics-based or empirical approaches. As a result, a wide range of data-driven and hybrid methodologies have been proposed to improve flood detection and mapping, particularly in data-rich environments\cite{amitrano2024flood}.
In recent years, artificial intelligence (AI) techniques have emerged as powerful tools for flood detection, benefiting from the growing availability of remote sensing data, in situ observations, and high-performance computing resources. Classical machine learning (ML) models, such as support vector machines, random forests, and gradient boosting, have been applied to classify flooded and non-flooded regions using multi-source datasets \cite{tanim2022flood, riazi2023enhancing}. More recently, deep learning (DL) architectures, including convolutional neural networks and recurrent neural networks, have demonstrated strong performance in extracting spatial and temporal patterns from satellite imagery \cite{stateczny2023optimized, wu2023near}. These approaches have significantly advanced the state of the art by improving detection accuracy, automating feature extraction, and enabling near-real-time flood monitoring.
Nevertheless, classical ML and DL methods face fundamental limitations when dealing with extremely high-dimensional feature spaces, complex nonlinear dependencies, and large-scale datasets. Issues such as computational cost, model generalization, and sensitivity to data imbalance and noise remain persistent challenges. Furthermore, the expressive power of classical models may be insufficient to capture subtle correlations in certain hydrological systems, particularly when the underlying dynamics are governed by multiscale interactions and rare extreme events. These challenges motivate the exploration of alternative computational paradigms that can potentially enhance learning efficiency and representational capacity.
In this context, quantum machine learning (QML) has recently emerged as a promising research direction at the intersection of quantum computing and AI. QML aims to leverage quantum mechanical principles, such as superposition, entanglement, and quantum interference to enhance classical learning algorithms\cite{biamonte2017quantum}. Within the QML framework, classical input features are mapped into quantum states using a quantum feature map, typically encoded into qubits or qudits\cite{huang2021power}. Quantum circuits then act on these states, transforming them into richer representations in high-dimensional Hilbert spaces. Measurements of the resulting quantum states are used to compute observables, kernels, or cost functions, which are subsequently optimized using classical routines.
This embedding of classical data into quantum Hilbert spaces constitutes a key advantage of QML, as it enables access to exponentially large feature spaces that are intractable for classical computers to represent explicitly\cite{schuld2015introduction}. As a result, QML models have the potential to uncover complex patterns and correlations that may be difficult to detect using conventional ML or DL techniques. From a theoretical perspective, QML offers opportunities for achieving quantum advantage in learning tasks, particularly in problems characterized by high dimensionality, strong nonlinearities, and limited training data.
Several classes of QML approaches have been proposed in the literature. Quantum kernel methods, which are currently the most mature and compatible with noisy intermediate-scale quantum (NISQ) hardware\cite{bharti2022noisy}, use quantum computers to evaluate similarity measures between data points in quantum feature space. Variational quantum circuits (VQCs) rely on parameterized quantum circuits whose parameters are trained through classical optimization\cite{benedetti2019parameterized}, enabling flexible and expressive models analogous to neural networks. A third category, quantum-enhanced feature extraction, embeds quantum components within classical ML pipelines to improve feature representation while retaining classical training frameworks\cite{cerezo2022challenges}.
Despite rapid methodological developments in QML, its application to geoscience and hydrology, particularly flood detection, remains largely unexplored. To date, most QML studies have focused on benchmark datasets or synthetic problems, with only limited attention given to real-world environmental systems. This gap highlights a significant opportunity to investigate whether quantum-enhanced learning models can offer tangible benefits for flood detection, especially in scenarios involving high-dimensional remote sensing data and complex hydrological interactions.

\begin{figure}[htbp]
\centering
\includegraphics[width=\columnwidth,height=0.95\textwidth,keepaspectratio]{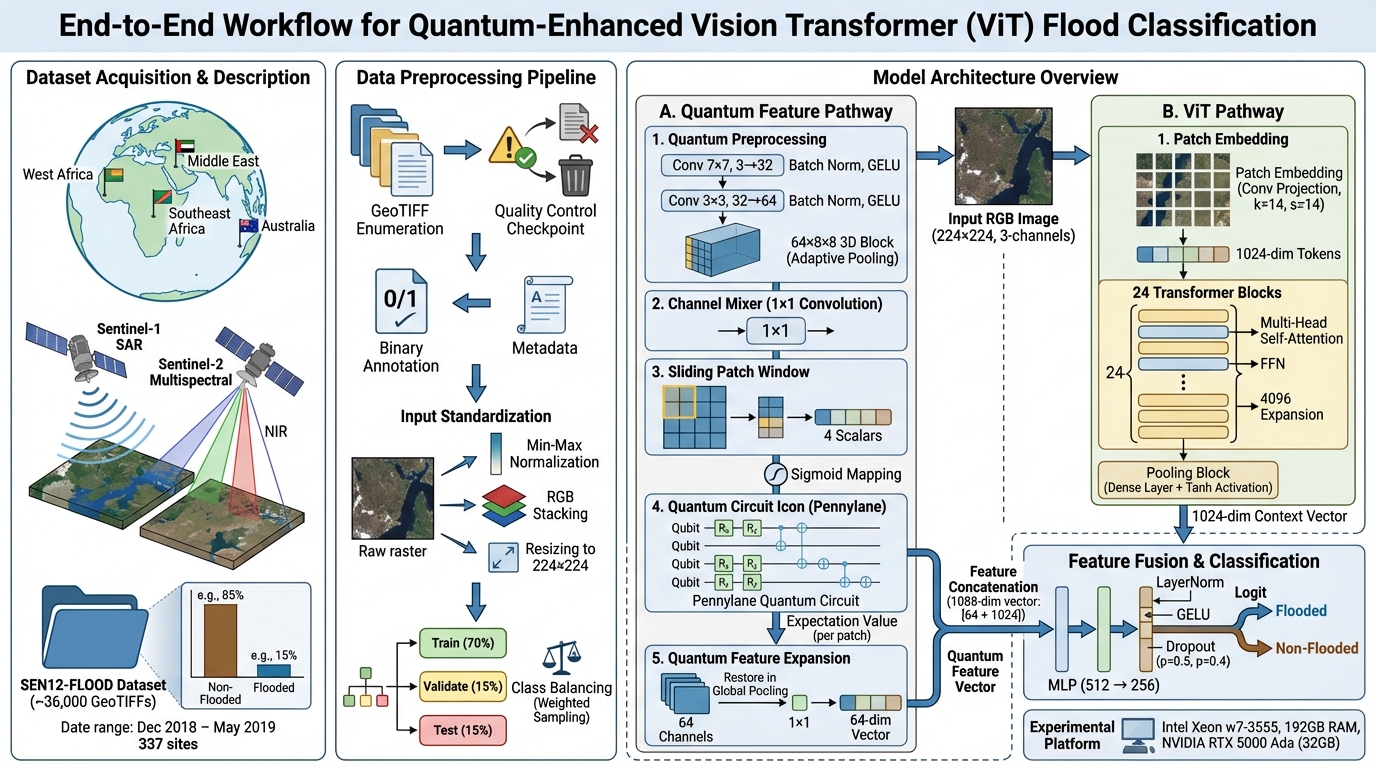}
\caption{The framework proposed in this study.}
\label{fig:workflow}
\end{figure}

\section{Methodology}

The proposed framework in this study was structured into four distinct phases, started with data acquisition followed by a rigorous pre-processing pipeline to standardize the inputs. Subsequently, we implemented a Quantum-enhanced ViT architecture to leverage hybrid feature extraction. In the final phase, the model was trained and its performance was compared to a classical ViT baseline to validate the efficacy of the quantum integration. Figure \ref{fig:workflow} provides an overview of the proposed experimental workflow. Both the proposed Quantum-enhanced ViT and the classical ViT base design use the same ViT backbone configuration, 1024 hidden dimensions, 24 transformer encoders, 16 focus heads, 4096 intermediate MLP size and patch size of $14\times14$, which gives approximately 304 million trainable parameters for the ViT alone. Crucially, both models were initialized with no pre-trained weights, which ensures that any performance differences between the two architectures can only be attributed to the structural contribution of the quantum-measurement segment and the fusion classifier, and not to the transfer of the learning benefits of the large-scale training. Full final fine-tuning was carried out on all layers of the transformer encoder for both models. No layer freezing or partial fine-tuning schedule was used, which allowed the focus masses to adapt from random start to the spectral and spatial characteristics of the SAR and multispectral imaging. All computations, including training and inference, were conducted on a high-performance workstation equipped with an Intel(R) Xeon(R) w7-3555 processor, 192 GB of RAM, and an NVIDIA RTX 5000 Ada Generation (32 GB) GPU.

\subsection{Dataset}

\begin{figure}[htbp]
\centering
\includegraphics[width=\columnwidth,height=0.35\textheight,keepaspectratio]{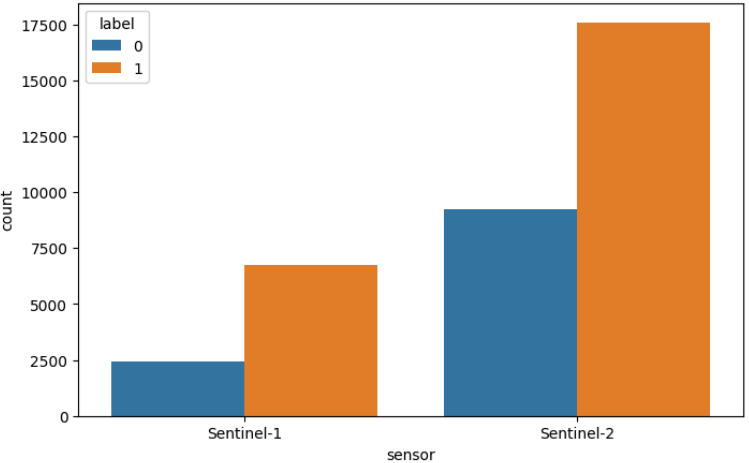}
\caption{Class distribution of SEN12-FlOOD dataset. Here label 0 refers to "Non-Flooded" class and label 1 refers to "Flooded" class.}
\label{fig:class_count}
\end{figure}

In this study, we used the publicly available SEN12-FLOOD dataset \cite{sen12flood}, comprising approximately 36,000 GeoTIFF files. Figure \ref{fig:class_count} shows the class distribution of the dataset. It contains co-registered multi-temporal imagery from Sentinel-1 synthetic aperture radar (SAR) and Sentinel-2 multispectral instruments, curated for supervised flood versus non-flood classification. Data were collected between December 2018 and May 2019 across 337 geographically distinct sites spanning West and Southeast Africa, the Middle East, and Australia, facilitating evaluation under heterogeneous hydro-climatic conditions and land-cover types. For each location, the dataset provides Sentinel-2 Level-2A surface reflectance time series with 12 spectral bands, alongside Sentinel-1 SAR time series in dual polarizations (VV, VH), processed using standard radiometric calibration and terrain correction workflows. This paired SAR-optical design enables robust flood mapping under cloud-covered conditions while retaining complementary spectral cues from multispectral observations.

\subsection{Dataset Preprocessing}

A data pipeline was designed to construct reliable flood annotations, filter unusable imagery, and standardize satellite tiles into a tensor\cite{kolda2009tensor} format suitable for deep learning. The process started by recursively enumerating all GeoTIFF files within the dataset directory and grouping them by parent folder, where each folder corresponded to a distinct geographic region. Ground truth labels were derived from the accompanying metadata, where a binary flood indicator was assigned to each region based on reported flooding events; all image tiles within a region inherited the parent folder's label. To handle metadata inconsistencies, edge cases where image folders lack corresponding metadata entries were flagged and assigned a conservative default label to ensure deterministic processing.

For the sake of data integrity, a quality control protocol was applied prior to training. The pipeline validated each file by attempting to read the first raster band, discarding corrupted or unreadable files. Additionally, tiles contained minimal information, characterized by near-zero variance or uniform pixel values, were filtered out to prevent the model from learning on blank or artifact-heavy data.

\begin{figure}[htbp]
\centering
\includegraphics[width=\columnwidth,height=0.35\textheight,keepaspectratio]{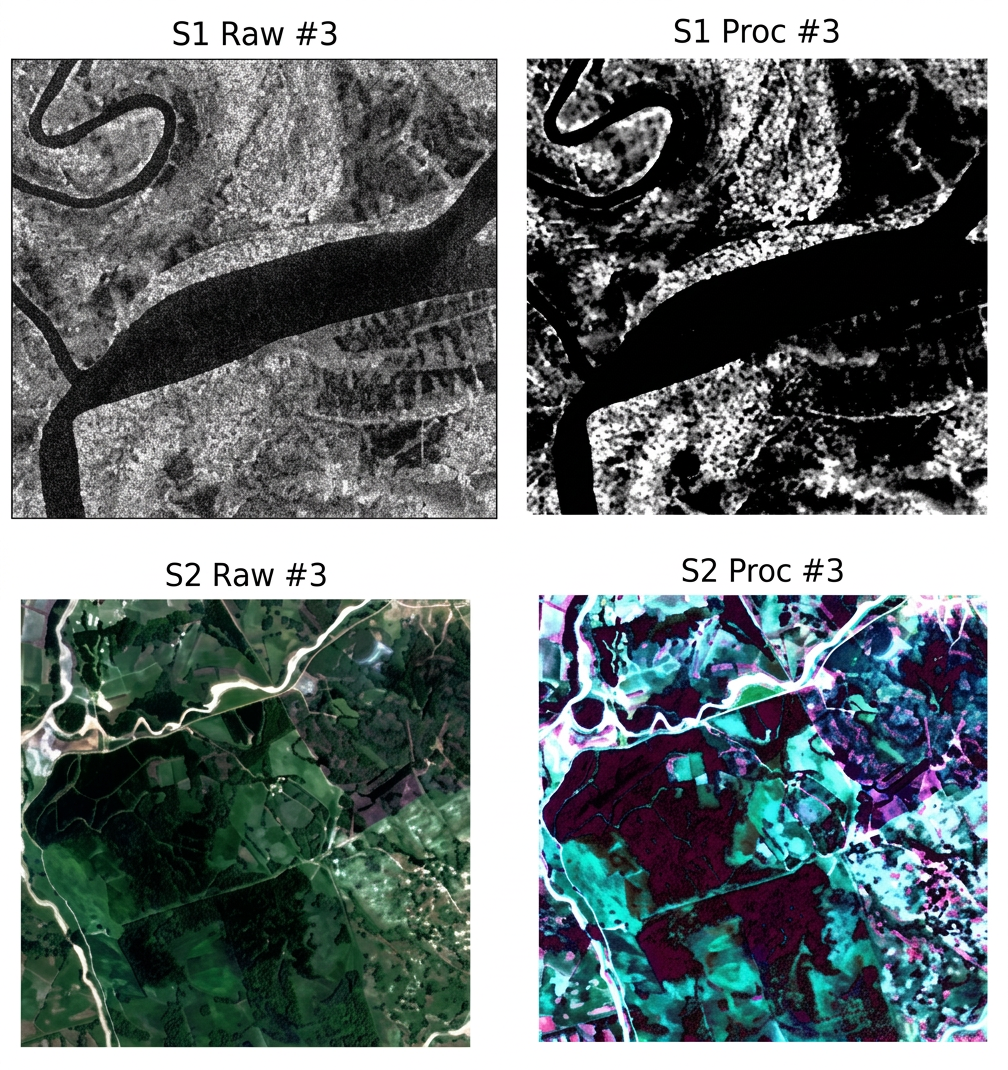}
\caption{GeoTIFF files (left) before and (right) after pre-processing.}
\label{fig:preprocessing}
\end{figure}

For model input standardization, each retained tile is converted into a vision-compatible format. Figure \ref{fig:preprocessing} presents how the primary raster band is min-max normalized to an 8-bit intensity image, replicated across 3-channels to form an RGB representation, and resized to 224 × 224 pixels. This channel replication is necessary to align the single-band spectral input with the standard 3-channel interface of the vision transformer backbone. The inputs are then normalized using ImageNet\cite{deng2009imagenet} and mean and standard deviation to stabilize optimization. During the training phase, data augmentation techniques are applied to improve generalization\cite{mikolajczyk2018data}. Validation and test sets undergo only resizing and normalization.

Finally, the curated dataset is partitioned into training 70\%, validation 15\%, and testing 15\% subsets. To mitigate class imbalance, we employ a weighted sampling strategy\cite{cui2019class} during training, where selection probabilities are computed based on inverse class frequencies. This ensures that mini-batches contain a balanced distribution of flood and non-flood samples, preventing the majority class from dominating the optimization process.

\begin{figure*}[htbp]
\centering
\includegraphics[width=\textwidth,height=0.35\textheight,keepaspectratio]{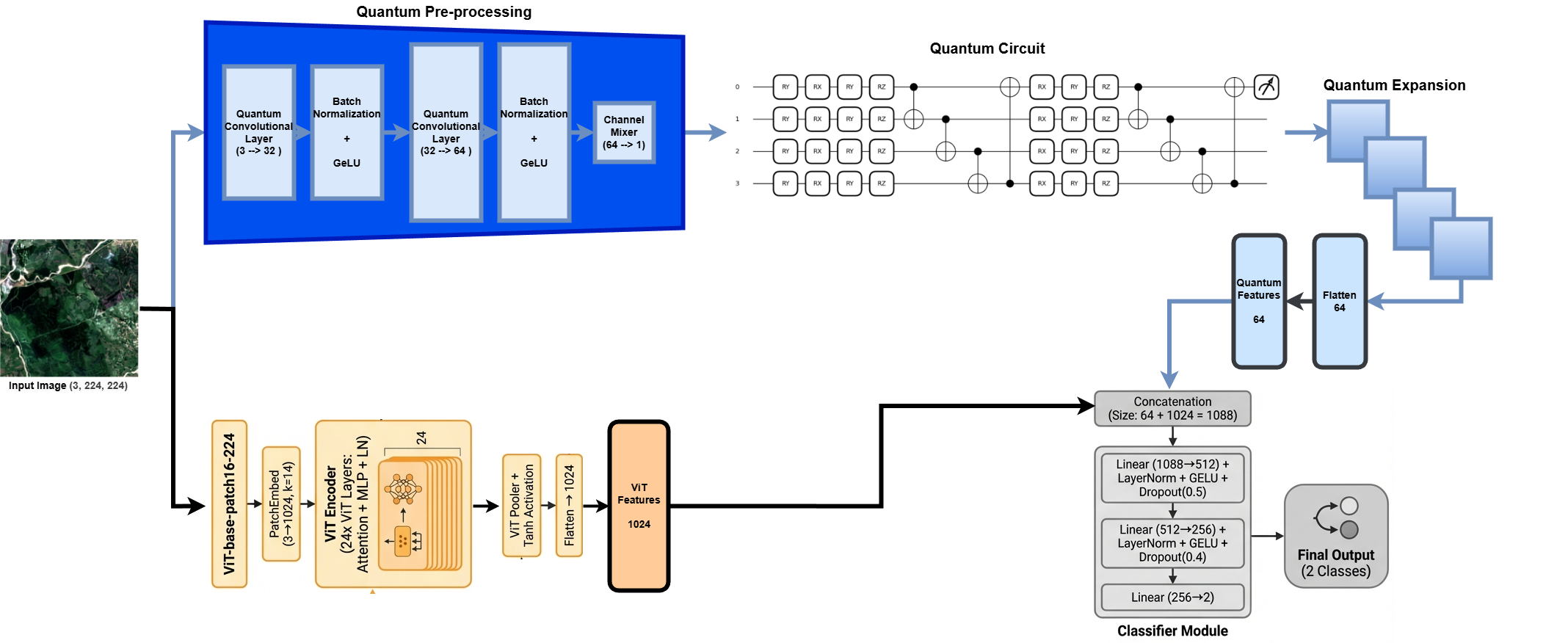}
\caption{Proposed quantum-enhanced vision transformer model architecture.}
\label{fig:model}
\end{figure*}

\subsection{Proposed Model Architecture}

In this study, we proposed a hybrid architecture by combining global visual context from a ViT\cite{dosovitskiy2020image} and localized quantum enhanced feature descriptors extracted via a parameterized quantum circuit. The network processes a 3-channel RGB input image through two parallel branches, a quantum feature pathway and a ViT pathway. Figure \ref{fig:model} shows the detailed overview of the proposed model architecture.

\subsubsection{Quantum Feature Pathway}

Quantum feature pathway consists of quantum preprocessing, quantum convolutional layer with 4-qubit parameterized quantum circuit using Pennylane\cite{bergholm2018pennylane} and quantum feature expansion. It begins with a preprocessing stage that uses two  quantum convolutional layers\cite{henderson2020quanvolutional} to downsample the image and expand its depth. The first layer uses large $7\times7$ kernels to map the 3 input channels into 32, while the second layer uses $3\times3$ kernels to increase it into 64. Both layers utilize batch normalization and GELU\cite{hendrycks2016gaussian} activation. An adaptive average pooling layer then condenses the feature map to a fixed grid of size $64\times8\times8$, providing a manageable input size for the quantum circuit. This output is passed through a channel mixer, which acts as a bottleneck by squeezing the 64 channels into a single channel using a $1\times1$ convolution.  A sliding window operation then extracts localized patches. Each $2\times2$ patch is flattened into a vector of 4 scalars, mapped to the range $[0,\pi]$ via a sigmoid function\cite{narayan1997generalized}, and encoded into a 4-qubit parameterized quantum circuit using rotation gates. The circuit expectation value is determined to produce a single scalar per patch, resulting in a spatially downsampled quantum feature map of size $1\times4\times4$. Finally, the block restores the 64 channels, applies global pooling to reduce the spatial size to $1\times1$, and flattens the result into a 64-dimensional quantum feature vector.

\subsubsection{ViT Pathway}

The ViT backbone\cite{wu2020visual} captures long-range dependencies by splitting the image into patches using a convolutional projection with a kernel size and stride of 14, embedding them into a 1024 dimensional tokens. A sequence of 24 transformer blocks refines the patch tokens through global self-attention followed by a position-wise feed-forward network. For the ViT, the feed-forward subnetwork expands the intermediate dimension to 4096. After passing through these layers and a final pooling stage, the token is processed by a dense layer and Tanh activation, producing a high-level context vector of size 1024.

In the final stage, the model performs feature fusion and classification. The 64-dimensional vector from the quantum stream is concatenated with the 1024-dimensional vector from the ViT stream to form a combined representation of size 1088. This fused vector is fed into the classifier, a sequential multi-layer perceptron\cite{rosenblatt1958perceptron}. The classifier progressively reduces the dimensionality through hidden layers of size 512 and 256. Each intermediate stage includes layer normalization for stability, GELU activation for nonlinearity, and dropout rates ($p=0.5$ and $p=0.4$) for regularization. Both the models were optimized using AdamW with a $1\times10^-6$ point learning rate, a weight decay of 0.05 to stabilize the training. The final layer outputs 2 unnormalized logits corresponding to the "Flooded" and "Non-Flooded" classes.

\section{Results and Discussion}

In this section, we compare the proposed quantum-enhanced ViT with the classical ViT baseline for binary flood detection from the satellite imagery. Both approaches followed the same training protocol to keep the comparison controlled and unbiased. For optimization, we applied the AdamW approach with an initial learning rate of $1\times10^{-4}$ and weight decay of $0.05$. We used a cosine-annealing\cite{loshchilov2016sgdr} learning-rate schedule with a warmup phase, the learning rate increases linearly during the first five epochs and then decays according to a cosine schedule. Model parameters were learned using CrossEntropyLoss\cite{mao2023cross} applied to the final fused representation. To assess the models performance, we calculated the accuracy, precision, recall, and F1-score from the confusion matrix as follows:

\begin{equation}
    \text{Accuracy} = \frac{TP + TN}{TP + TN + FP + FN}
    \label{eq:accuracy}
\end{equation}

\begin{equation}
    \text{Precision} = \frac{TP}{TP + FP}
    \label{eq:precision}
\end{equation}

\begin{equation}
    \text{Recall} = \frac{TP}{TP + FN}
    \label{eq:recall}
\end{equation}

\begin{equation}
    \text{F1-score} = 2 \times \frac{\text{Precision} \times \text{Recall}}{\text{Precision} + \text{Recall}}
    \label{eq:f1score}
\end{equation}

where $TP$, $TN$, $FP$, and $FN$ denote the true positives, true negatives, false positives, and false negatives, respectively.

\begin{table}[htbp]
\caption{The accuracy and F1-score values calculated for the classical ViT baseline and quantum-enhanced ViT models.}
\label{tab:result-1}
\centering
\begin{tabular}{lcc}
\hline
\textbf{Metric} & \textbf{Classical ViT baseline} & \textbf{Quantum-enhanced ViT model} \\
\hline
Accuracy & 84.48\% & \textbf{94.47\%} \\
F1-Score & 0.841 & \textbf{0.944} \\
\hline
\end{tabular}
\end{table}

Table \ref{tab:result-1} summarizes the overall test performance for the two architectures. The classical ViT baseline achieves 84.48\% accuracy with an F1-score of 0.841, which leaves limited margin for a safety-critical application such as flood mapping. After adding the quantum-enhanced feature pathway and fusing it with the ViT representation, both aggregate metrics increase substantially to 94.47\% for the accuracy and 0.944 for the F1-score.  

\begin{table*}[htbp]
\caption{Performance Comparison in Terms of Per-Class Evaluation.}
\label{tab:result-all}
\centering
\begin{tabular}{lcccccc}
\hline
\multirow{2}{*}{\textbf{Classes}} & \multicolumn{3}{c}{\textbf{Classical ViT Baseline}} & \multicolumn{3}{c}{\textbf{Quantum-Enhanced ViT Model}} \\
\cline{2-4} \cline{5-7}
 & \textbf{Precision} & \textbf{Recall} & \textbf{F1-Score} & \textbf{Precision} & \textbf{Recall} & \textbf{F1-Score} \\
\hline
Flooded & 85.61\% & 92.42\% & 0.8888 & \textbf{94.18\%} & \textbf{97.81\%} & \textbf{0.9596} \\
Non-Flooded & 81.49\% & 68.24\% & 0.7428 & \textbf{95.14\%} & \textbf{87.65\%} & \textbf{0.9124} \\
\hline
\end{tabular}
\end{table*}

Table \ref{tab:result-all} provides a class-wise breakdown that helps explain the overall performance gains. For the classical ViT baseline, the flooded class recall is relatively high (92.42\%), whereas performance is noticeably weaker for the non-flooded class, particularly in terms of recall (68.24\%), indicating frequent confusion in rejecting false flood detections. In contrast, the proposed hybrid model achieves higher precision and recall for both classes, with the most pronounced improvement observed for the non-flooded class: recall increases to 87.65\% and precision to 95.14\%. These improvements are consistent with a substantial reduction in false positives over non-flooded regions and suggest that the additional quantum-derived descriptors provide complementary information that the transformer-only baseline does not capture as effectively.

\begin{figure}[!t]
\centering
\includegraphics[width=\columnwidth,height=0.35\textheight,keepaspectratio]{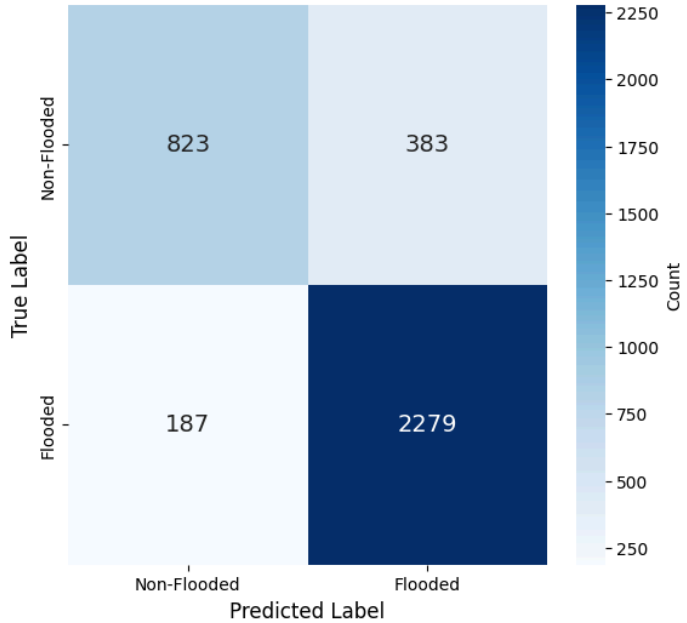}
\caption{The confusion matrix for the classical ViT baseline.}
\label{fig:cf-vit}
\end{figure}

\begin{figure}[!t]
\centering
\includegraphics[width=\columnwidth,height=0.35\textheight,keepaspectratio]{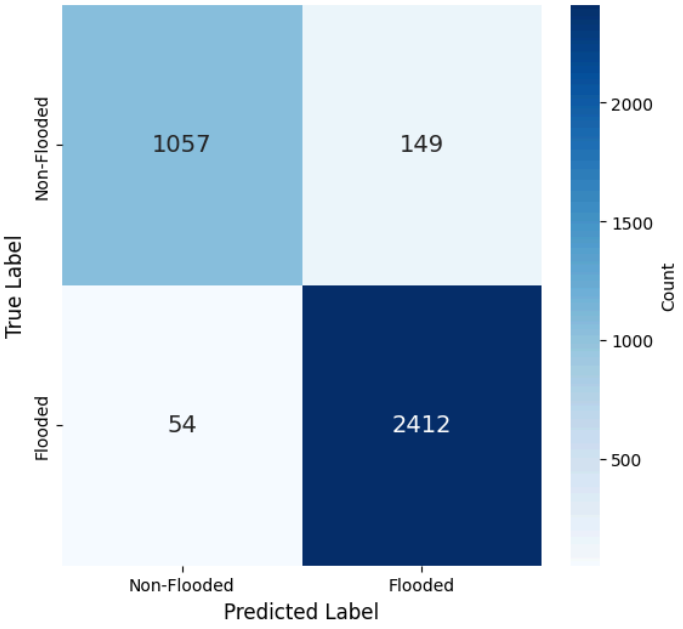}
\caption{The confusion matrix for the quantum-enhanced ViT model.}
\label{fig:cf-qvit}
\end{figure}

The confusion matrices in Figures \ref{fig:cf-vit} and \ref{fig:cf-qvit} visually corroborate the quantitative results. Compared with the classical baseline, the proposed model exhibits fewer misclassifications on the held-out test set, particularly for the minority non-flooded class, consistent with the higher recall and precision reported in Table \ref{tab:result-all}. Overall, these findings indicate that incorporating the quantum feature branch enhances the learned representation and improves discrimination between flooded and non-flooded patterns, especially in challenging scenes.

\section{Conclusion}

This study presented a novel hybrid deep learning architecture designed to address the critical challenge of flood detection in satellite imagery. By integrating a parameterized quantum convolution layer with a vision transformer backbone, we leveraged the strengths of both paradigms: the global contextual modeling capability of transformers and the enhanced feature extraction capacity of quantum circuits. Results showed that the proposed hybrid approach yielded consistent performance gains over a purely classical baseline. Most notably, the proposed model substantially reduced false positives in non-flooded regions, led to a marked increase in precision for the Non-Flooded class. This improvement indicated that the quantum-enhanced features provided a richer and more discriminative representation of complex terrain textures that often challenge traditional models. Overall, these findings suggest that quantum feature extraction can effectively complement transformer-based architectures by supplying compact, nonlinear descriptors that enhance discriminative power in geospatial classification tasks involving heterogeneous and visually complex imagery. While classical noise in SAR images has been partially mitigated by median filtering and percentile stretching, the effect of residual noise on the learned representations of quantum features is not quantified. The current implementation is based on the PennyLane noiseless qubit simulator; the evaluation of robustness in realistic quantum noise models, including depolarization errors and gate bias, is reserved for future work.

Future work will also focus on scaling the quantum component to support a larger number of qubits and deeper parameterized circuits, investigating alternative quantum encoding schemes beyond simple rotation gates, and extending the proposed framework to multi-temporal and multi-sensor flood datasets in order to assess robustness across diverse flooding scenarios and sensor modalities. In addition, optimizing the quantum circuit execution pipeline and exploring hybrid training strategies that reduce wall-clock time while preserving accuracy gains remain important directions toward making quantum–classical models more practical for large-scale remote sensing applications. 

\section{Acknowledgment}
BG is grateful to the University of Texas at Arlington for financial supports through faculty start-up fund and the STARs award.

\bibliographystyle{IEEEtran}
\bibliography{ref}
\vspace{12pt}

\end{document}